\documentclass[letterpaper, 10 pt, journal]{ieeeconf}  

\IEEEoverridecommandlockouts                              

\makeatletter
\newif\if@restonecol
\makeatother

\usepackage{cite}
\usepackage{amsmath,amssymb,amsfonts}
\usepackage{algorithmic}
\usepackage{graphicx}
\usepackage{textcomp}
\usepackage{xcolor}
\usepackage{booktabs}
\usepackage{bbding}
\usepackage{tikz}
\usepackage{bm}
\usepackage[ruled,linesnumbered,noline]{algorithm2e}
\usepackage{array}
\usepackage{makecell} 

\title{\LARGE \bf
Self-Reconfiguration Planning for Deformable Quadrilateral Modular Robots
}

\author{Jie Gu$^{1}$, Hongrun Gao$^{1}$, Zhihao Xia$^{1}$, Yirun Sun$^{1}$, Chunxu Tian$^{1,\dag}$, Dan Zhang$^{4,\dag}$ 
\thanks{*This work was supported by the National Nature Science Foundation of China (grants 52305012).}
\thanks{$^{1}$Institute of AI and Robotics, Academy for Engineering \& Technology, Fudan University, Shanghai 200433, China.}
\thanks{$^{2}$Department of Mechanical Engineering, The Hong Kong Polytechnic University, Hung Hom.}%
\thanks{$^{\dag}$Corresponding author is Chunxu Tian, email: chxtian@fudan.edu.cn and Dan Zhang, email: dan.zhang@polyu.edu.hk}
}

\begin{document}

\maketitle

\thispagestyle{empty}
\pagestyle{empty}

\begin{abstract}
For lattice modular self-reconfigurable robots (MSRRs), maintaining stable connections during reconfiguration is crucial for physical feasibility and deployability. This letter presents a novel self-reconfiguration planning algorithm for deformable quadrilateral MSRRs that guarantees stable connection. The method first constructs feasible connect/disconnect actions using a virtual graph representation, and then organizes these actions into a valid execution sequence through a Dependence-based Reverse Tree (DRTree) that resolves interdependencies. We also prove that reconfiguration sequences satisfying motion characteristics exist for any pair of configurations with seven or more modules (excluding linear topologies). Finally, comparisons with a modified BiRRT algorithm highlight the superior efficiency and stability of our approach, while deployment on a physical robotic platform confirms its practical feasibility.

\end{abstract}

\section{Introduction}
Modular self-reconfigurable robots (MSRRs) refer to robotic systems composed of homogeneous modules capable of altering their interconnections to dynamically reconfigure their morphology, thereby enhancing their adaptability to diverse environments. Existing MSRRs can be classified into five primary categories: lattice-type\cite{flyingmodularsaldana2018modquad}, chain-type\cite{yim2000polybot}, hybrid-type\cite{kurokawa2008M-TRANIII}, truss-type\cite{qin2022trussbot}, and free-form systems\cite{tu2022freesn,zhao2024snail-nature}. Lattice MSRRs are characterized by polygonal or polyhedral architectures, in which modules interconnect through edges or faces to form planar tessellations or spatially dense packings. In planar lattice MSRRs, triangular\cite{belke2017mori,pieber2018adaptive}, quadrilateral\cite{romanishin2013M-blocks}, hexagonal\cite{metamorphic_definition}, and circular geometries\cite{saintyves2022granulobot} are most common, whereas cubic\cite{romanishin20153d} and spherical\cite{liang2020freebot} configurations dominate in three-dimensional systems. This letter focuses on the quadrilateral MSRRs.

Self-reconfiguration in MSRRs refers to the process in which modules alter their topological connections, enabling the system to transform from an initial configuration to a target configuration. The corresponding self-reconfiguration planning algorithm determines a sequence of actions that achieves this transformation. Different structural designs exhibit distinct primitive motions for self-reconfiguration, and accordingly, various reconfiguration motion algorithms have been developed.

For quadrilateral MSRRs, self-reconfiguration can be classified into three categories: sliding, relocation, and pivoting. Sliding is rarely adopted in practical systems because of the challenges associated with modularizing guide rails and implementing linear actuation. As a result, most related algorithms focus on theoretical studies of abstract square modules\cite{abel2024sliding}.
Relocation refers to a motion in which a module detaches from the system and reattaches elsewhere. Such self-reconfiguration strategies and their corresponding planning algorithms are predominantly employed in mobile MSRRs\cite{liu2019distributed}, supported by a comprehensive research framework that focuses on optimality and efficiency. However, such reconfiguration inevitably introduces docking challenges, since detached modules must perceive and estimate their own positions independently.
Pivoting is a reliable and fundamental reconfiguration motion for MSRRs, characterized by rotation about a fixed pivot. However, existing algorithms for pivoting-based self-reconfiguration also remain largely theoretical \cite{akitaya2021universal}, with limited validation on physical platforms. This may be attributed to inherent weaknesses in the pivoting mechanism, as two modules have only a point-connection during the motion, making the connection weaker and more susceptible to accidental detachment. Such instability can lead to reconfiguration failure or even systemic collapse. Moreover, conventional pivoting inherently sweeps through adjacent cells, which must therefore remain unoccupied, imposing additional spatial constraints on the reconfiguration process.

Compared with MSRRs composed of rigid units, deformable MSRRs offer greater morphological flexibility\cite{piranda2021datom,modur,zhao2025tensegrity}. By integrating deformation into the pivoting process, they overcome several limitations of conventional pivoting, including improved connection stability and reduced spatial interference. Nevertheless, the novelty of their reconfiguration primitive motion prevents many existing algorithms from being directly adapted or deployed to such systems. 

This work presents the first self-reconfiguration planning algorithm tailored to deformable quadrilateral MSRRs, capable of generating a valid reconfiguration sequence whenever two configurations are theoretically reconfigurable.
The algorithm is designed around a new motion primitive, termed morphpivoting, which integrates controlled deformation into pivoting to enhance stability and eliminate geometric interference (formally defined in Section II).
First, we perform a configuration-space search based on a representation termed the virtual graph, which efficiently constructs a comprehensive set of feasible actions while preserving structural validity.
Second, we introduce Dependence-based Reverse Tree (DRTree), a novel ordering framework that organizes all actions into an executable sequence.
DRTree establishes a dependency-consistent hierarchical structure that resolves action conflicts, guarantees geometric feasibility throughout the morphpivoting process, and produces a deployable reconfiguration sequence suitable for execution on the physical robotic platform.

The main contributions of this letter are as follows:
\begin{enumerate}
    \item We propose a novel algorithm for planar quadrilateral MSRRs. The method consists of (i) a search method using a virtual graph to generate a comprehensive set of feasible actions, and (ii) a framework that organizes these actions into a conflict-free, deployable sequence executable on physical robots.
    \item We prove that configurations composed of multiple modules ($ \geq7 $) are isotypic (except linear configurations), ensuring that a reconfiguration sequence exists between any two theoretically reconfigurable configurations.
    \item The proposed algorithms are demonstrated on a quadrilateral modular robotic platform, and their efficiency and stability are validated through comparisons with baseline methods.
\end{enumerate}

The remaining content of this letter is organized as follows. Section II briefly introduces the hardware characteristics and motion characteristics of the robots. Section III presents the basic definitions and formally states the research problem. The self-reconfiguration planning algorithm is described in Section IV. Experimental results are provided in Section V. Section VI further supplements the theoretical proof of reconfigurability. Finally, Section VI concludes the letter and discusses directions for future work.

\section{Hardware Platform}
\begin{figure}[tbp]
\centerline{\includegraphics[width=\linewidth]{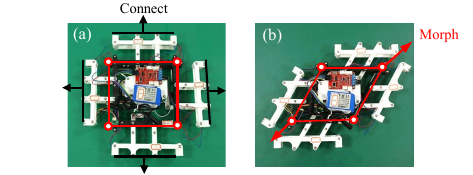}}
\caption{The hardware platform. The standard square module with four sides capable of connecting to other modules. (b) The rhombus shape obtained by morphing the module in (a) along its diagonal.}
\label{fig:rhombot}
\end{figure}

Our algorithmic deployment platform is a deformable MSRR with a rhombus-based geometry, as shown in Fig.~\ref{fig:rhombot}. Unlike conventional rigid square MSRRs, this module can fold along its diagonal axes, thereby achieving morphing. This deformability provides a foundation for reconfiguration with each module maintaining edge connections with the overall system.

\begin{figure}[tbp]
\centerline{\includegraphics[width=\linewidth]{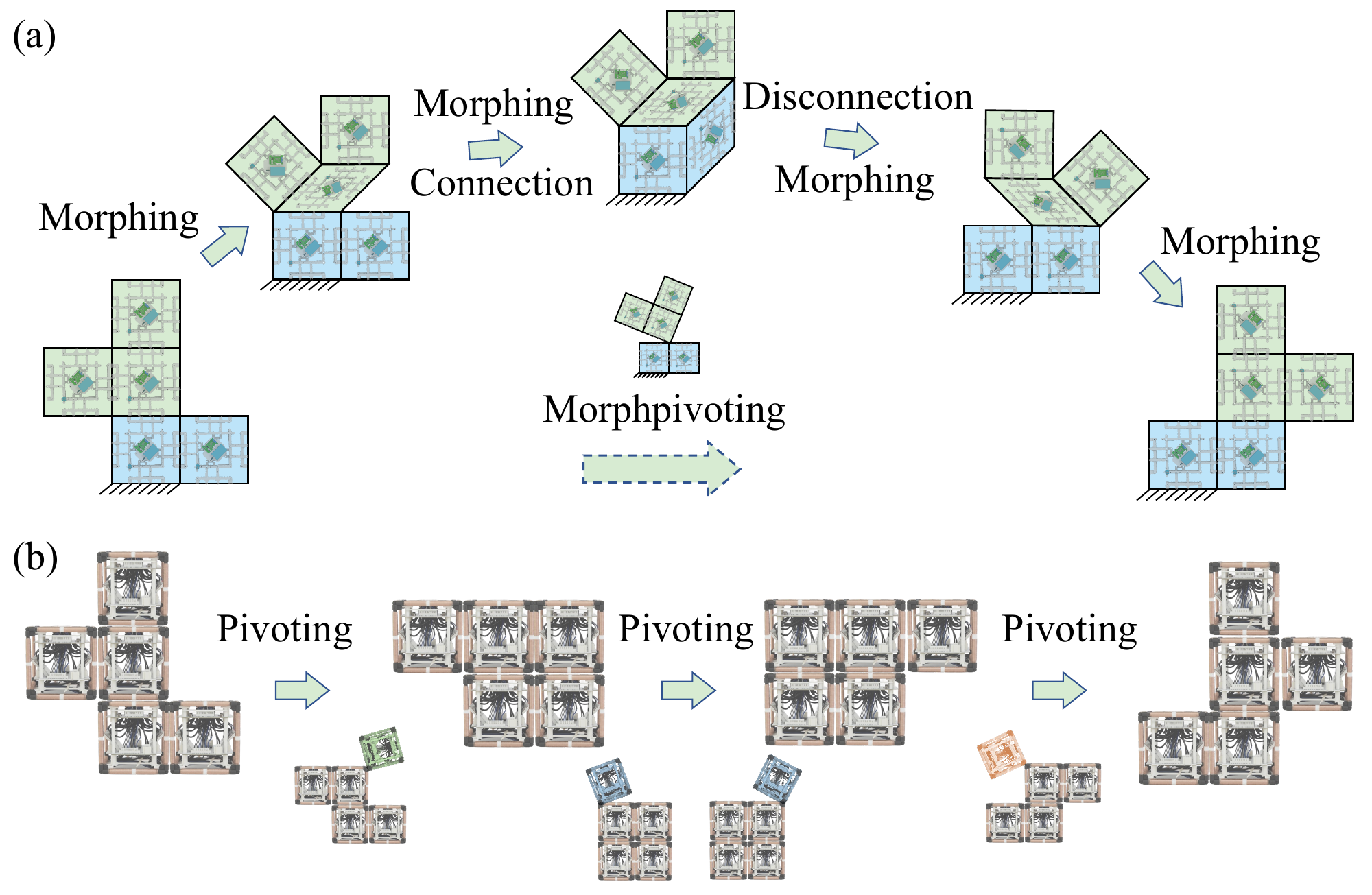}}
\caption{Morphpivoting: definition and comparison with conventional pivoting. (a) The proposed quadrilateral MSRR model achieves reconfiguration through morphing combined with necessary connection and disconnection operations. (b) Reconfiguration via pivoting, demonstrated using the ElectroVoxel system.}
\label{fig:morphpivoting}
\end{figure}

In contrast to traditional pivoting used in rigid MSRRs, such as the ElectroVoxel\cite{nisser2022electrovoxel}, where reconfiguration is achieved through a single rotation along the module’s vertex, our platform realizes reconfiguration through a sequence of actions composed of morphing, connection, and disconnection, as illustrated in Fig.~\ref{fig:morphpivoting}.

This sequence of sub-actions can be functionally regarded as an equivalent primitive motion, which we term morphpivoting. Morphpivoting treats these physically continuous sub-actions as a unified motion for analysis and control, and exhibits the following key characteristics:
\begin{itemize}
    \item During all sub-actions, every module remains connected to the overall system through edge-based attachments, ensuring structural stability and preventing accidental detachment.
    \item Morphpivoting can involve not only a single module but also a group of interconnected modules acting as a collective unit.
    \item At least three modules must share a pivot point for the morphpivoting process to be feasible.
\end{itemize}

\section{Definition and Problem Formation}
Although our robot possesses the capability to transform into rhombic shapes, we designate the square geometry as its canonical configuration. This constraint simplifies system initialization, while rhombic deformations are permitted only during reconfiguration. Consequently, the canonical configuration of our abstract model is initially defined by a finite set of squares, and the collection of all configurations containing $n$ squares is formally denoted as $\mathcal{C}_n$. The system is embedded in a Cartesian coordinate frame to enable precise positional specification of each square. Two squares are considered connected if the distance between them equals one unit.

Each square possesses four edges that can potentially engage in a connection. 
When a square is isolated and not attached to the main system, all four edges are geometrically isomorphic owing to its rotational symmetry. 
However, once one edge of the square is connected to the robotic structure, this symmetry is broken, and the remaining three edges become non-isomorphic. 
Specifically, the directions of the adjacent edges intersect with the current connection, whereas the direction of the opposite edge is collinear with it. 
Therefore, the edges and their orientations must be regarded as distinct entities. The four edges of a square are defined by their outward normal directions, denoted as $Dir$, as follows:
\begin{equation}
\begin{aligned}
Dir = \{ 
&Top:(0,1);Left:(1,0);\\
&Bottom:(0,-1);Right:(-1,0)
\}.
\end{aligned}
\label{eq:dir_def}
\end{equation}

When establishing a new connection, the relative geometric positions of the two participating squares are used to determine which $Dir$ values are involved in the connection. 
After the connection is formed, the corresponding $Dir$ is replaced by the new connection.

\begin{figure}[tbp]
\centerline{\includegraphics[width=\linewidth]{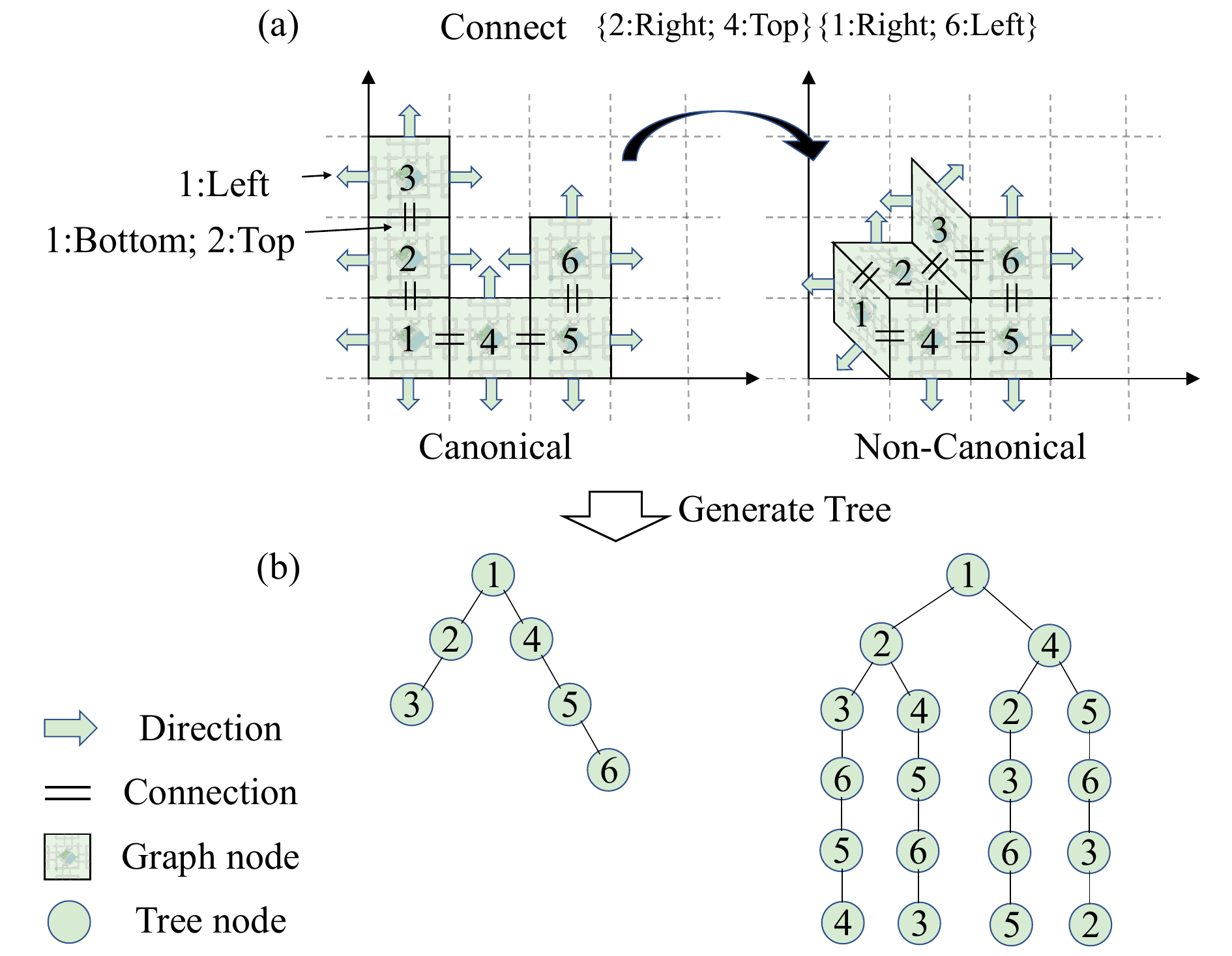}}
\caption{(a) The canonical and non-canonical forms of a six-module configuration graph. The latter is obtained from the former by adding $e_{24}$ and $e_{36}$. (b) The tree generated from the two forms in (a).}
\label{fig:Configuration_Graph}
\end{figure}

A graph is an effective representation method for describing the configuration of MSRR, as it clearly illustrates the connections between modules. The connection graph of a configuration is defined as \( G = (V, E) \), where \( V \) represents the modules and \( E \) represents the connections between modules. For example, the connection between $v_i \in V$ with $Dir1$ and $v_j \in V$ with $Dir2$ is denoted as
\(e_{ij\{Dir1,Dir2\}} \in E = (v_i: Dir1; v_j: Dir2 \mid i, j \in \{1, 2, \ldots, n\},\, i \neq j)\)
and is abbreviated as $e_{ij}$ since typically there exists only one connection between two modules.
A six-module configuration graph in $\mathcal{C}_6$ is illustrated in Fig.~\ref{fig:Configuration_Graph}(a) with its non-canonical form achieved by adding $e_{24}$ and $e_{36}$.

A graph is called a tree if it is connected and contains exactly one simple path between any pair of vertices, which requires the designation of a root. The tree is then expanded according to the current connection relationships. In this way, new connections are typically formed between derivative vertices that reside at equivalent hierarchical levels within the tree. The tree generated from Fig.~\ref{fig:Configuration_Graph}(a) is illustrated in Fig.~\ref{fig:Configuration_Graph}(b). When loops are present in the graph, the corresponding tree is extended, as previously visited vertices are allowed to be revisited during expansion, except for those lying on the path to the root.

Based on the aforementioned definitions, the core problem can be formally stated as follows: Given an initial configuration $G_1 =(V_1,E_1)$ and a goal configuration $G_2 =(V_2,E_2)$, if they are mutually transformable, that is, \textit{\textbf{isotypic}}, the objective is to find a topological reconfiguration sequence.
\begin{equation}
     \mathcal{TRC}(G_1,G_2) = \{s_m\circ ... \circ s_2 \circ s_1\} 
\end{equation}
where $s_i,i \in \{1,2,...,m\}$ is a morphpivoting and the symbol \( \circ \) denotes the nesting of actions, indicating that actions are executed sequentially from \( s_1 \) to \( s_m \).  

\section{Reconfiguration Planning Algorithm}
\subsection{Virtual Graph}

A configuration may contain multiple loops or possess the potential to form new ones. Loop generation and decomposition serve as the fundamental mechanisms underlying configuration transformations. While loop decomposition is generally unconstrained, provided that no module becomes isolated, loop generation is subject to three constraints imposed by the morphpivoting motion characteristics:

\begin{itemize}
    \item[] \textbf{Constraint 1:} The number of vertices in a loop must satisfy $|V| \geq 3$.
    \item[] \textbf{Constraint 2:} No unconnected directions are permitted within the newly formed loop.
    \item[] \textbf{Constraint 3:} Two distinct loops cannot share two or more vertices.
\end{itemize}

\begin{figure}[tbp]
\centerline{\includegraphics[width=\linewidth]{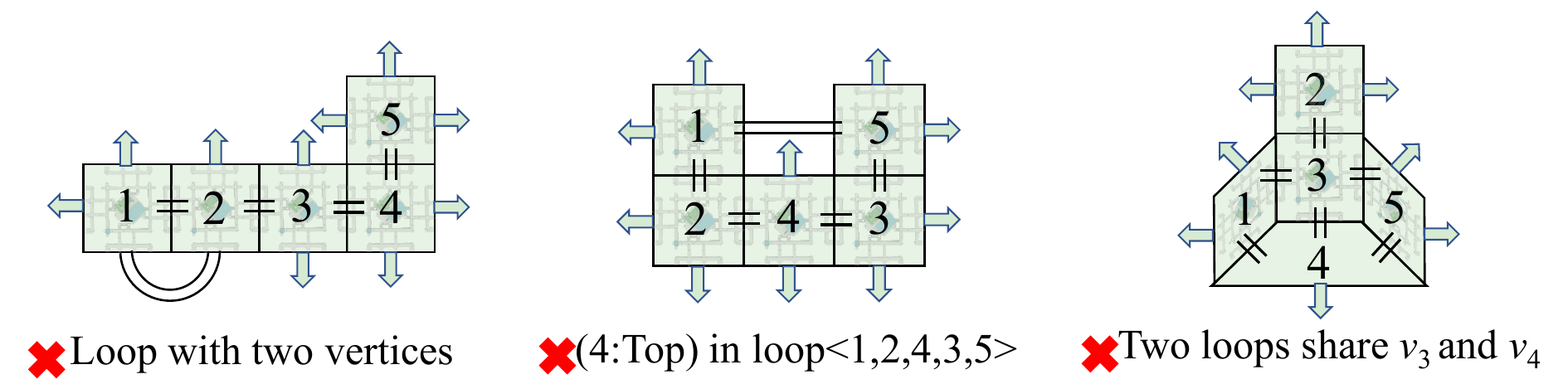}}
\caption{Three counterexamples that violate the above constraints.
}
\label{fig:constraint}
\end{figure}

\begin{figure}[tbp]
\centerline{\includegraphics[width=\linewidth]{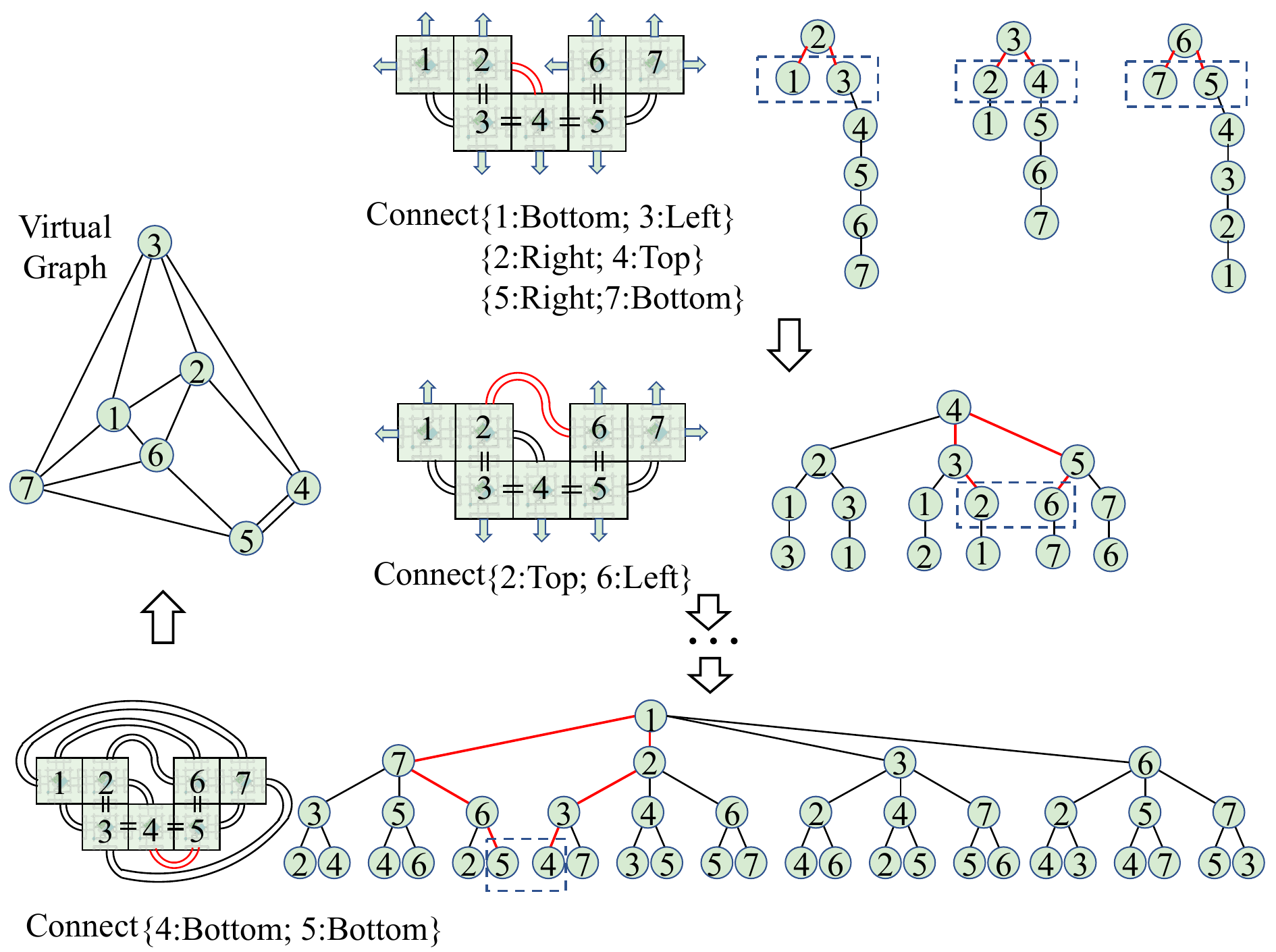}}
\caption{Illustration of the virtual graph generation process.}

\label{fig:virtual_graph}
\end{figure}

\begin{figure*}[tbp]
\centerline{\includegraphics[width=\linewidth]{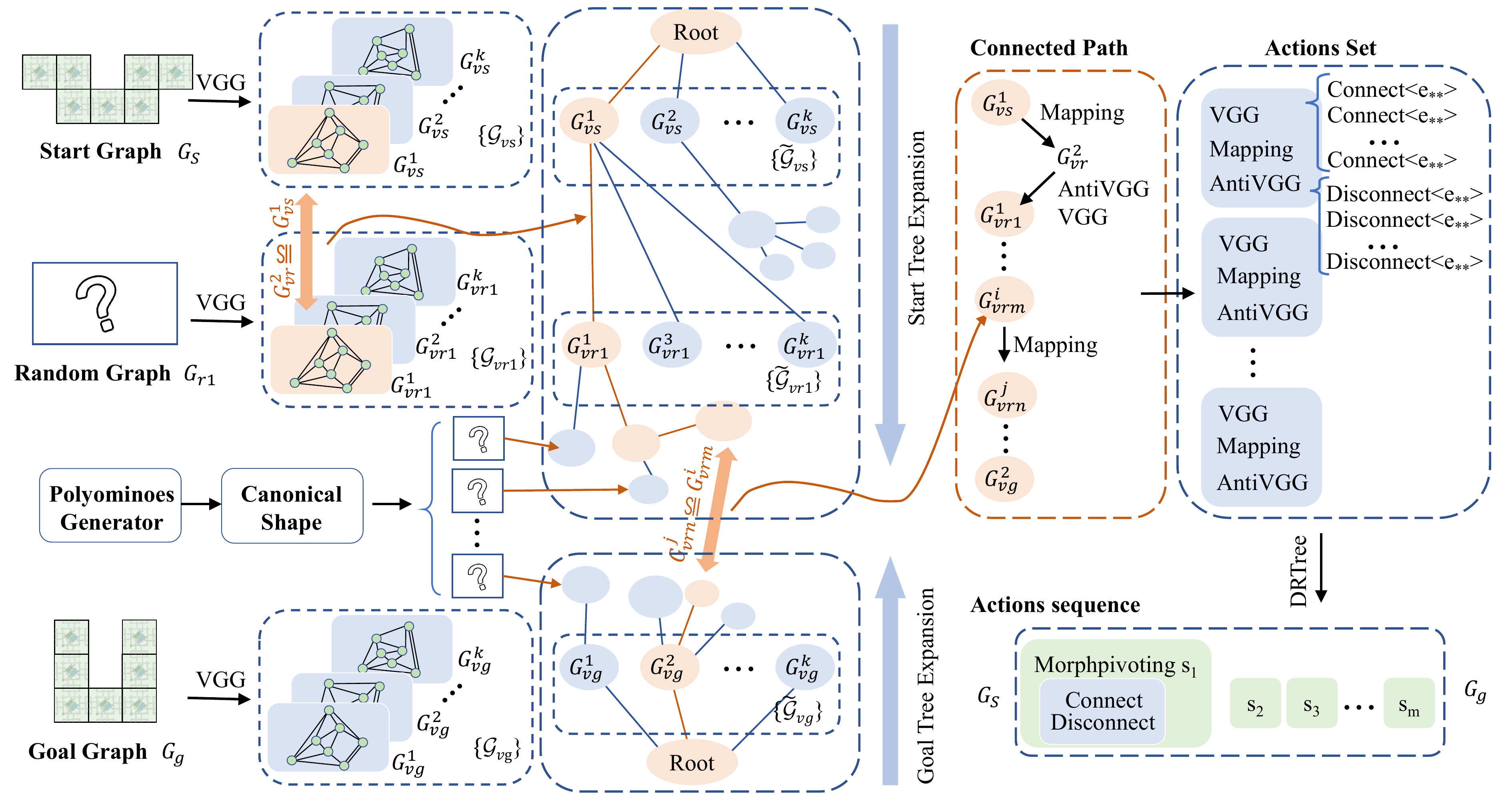}}
\caption{Schematic diagram of the proposed reconfiguration planning algorithm.}
\label{fig:reconfiguration}
\end{figure*}

Three counterexamples that violate the above constraints are depicted in Fig.~\ref{fig:constraint}. 
Regarding Constraint~3, although a graph containing two loops that share two vertices is geometrically infeasible due to conflicting edges, 
a valid configuration can still be obtained from such an invalid graph by selectively removing certain edges from the conflicting edge set. 
This observation implies that graphs violating Constraint~3 may still contain meaningful reconfiguration pathways once the conflicting edges are selectively removed. 
Therefore, in order to comprehensively explore all potential topological transitions, 
we disregard Constraint~3 while preserving Constraints~1 and~2. 
Under this relaxation, all possible virtual edges are introduced into the original graph, 
resulting in a new graph termed the \textit{virtual graph}, denoted as $G_v$.

The generation principle of the virtual graph is illustrated in Fig.~\ref{fig:virtual_graph}. 
At each stage, every vertex \( v \in V \) is treated as a root node to construct a hierarchical tree, 
following the procedure described in Section~III. 
Within the same hierarchical level of the tree, pairs of vertices that satisfy Constraints~1 and~2 are connected to establish new edges. 
Subsequently, the tree is updated. 
The above process is iteratively repeated until no further connections can be established. 
It is worth noting that the final connection established between vertices~4 and~5 in the last stage does not violate Constraint~1, 
because vertices~4 and~5 are both third-generation descendants of vertex~1 within the hierarchical tree. 
This connection forms part of the loop \(\langle 4,3,2,1,7,6,5 \rangle\).
Once the virtual graph is fully constructed, we no longer focus on the orientations of the modules, resulting in a topological graph.

Specifically, a virtual graph generation algorithm (VGG) is detailed in Algorithm~1. In lines 1–4, each module is placed within a Cartesian coordinate system and assigned its physical position. Based on this information, an initial graph is constructed and designated as the root node of a tree. Each subsequent call to the \texttt{MakeValidConnect} function expands this tree, and the resulting leaf nodes correspond to the generated virtual graphs. In lines 5–15, all potential connection pairs are exhaustively evaluated for validity, yielding non-redundant valid pairs while eliminating duplicates. In lines 16–22, conflicting pairs are detected to resolve topological bifurcations. Divergent virtual graphs with distinct topologies are preserved as child nodes of the current virtual graph, thereby enabling recursive tree expansion through conflict resolution.

During the virtual graph generation process, a $Dir$ of a vertex may encounter multiple directional connection options, where each selection distinctively affects subsequent virtual connections. This phenomenon enables a single configuration to generate multiple topologically non-isomorphic virtual graphs. As illustrated in Fig.~\ref{fig:reconfiguration}, the VGG transforms the start graph $G_s$ into multiple virtual graphs denoted as $G_{vs}^1$, $G_{vs}^2$, \ldots, $G_{vs}^k$, whose collection is represented by $\mathcal{G}_{vs}$. 

\subsection{Bidirectional Isomorphism Tree}
For two given configurations $G_{s}$ and $G_{g}$, their corresponding sets of virtual graphs are denoted as $\mathcal{G}_{vs}$ and $\mathcal{G}_{vg}$, respectively. In the simplest case, if there exists a pair of isomorphic members between these two sets, the corresponding virtual graphs can be transformed into each other through a relabeling action of the graph vertices, termed as \textit{mapping}. In this letter, the mapping is implemented using the VF2 graph isomorphism algorithm\cite{cordella2004VF2}. Consequently, the topological reconfiguration sequence from configuration $G_{s}$ to $G_{g}$ can be formulated as:
\begin{equation}
   \mathcal{TRC}(G_{s}, G_{g}) = \{\text{VGG}, \text{Mapping}, \text{AntiVGG}\}.
   \label{eq3}
\end{equation}

If no such isomorphic pair is found, there must exist a sequence of intermediate configurations that serve as bridges according to \textbf{Corollary 1}, discussed in Section VI. Therefore, by repeatedly applying Eq.~(\ref{eq3}) until the target configuration is reached, the entire process can be modeled as:

\begin{equation}
\begin{split}
   & \mathcal{TRC}(G_s, G_{g}) \\
   & = \mathcal{TRC}(G_{rk}, G_{g})  \circ ...\circ \mathcal{TRC}(G_{r1}, G_{r2}) \\
   &\circ \mathcal{TRC}(G_s, G_{r1})
\end{split}
\label{eq4}
\end{equation}

where, \( k=\mathbb{N}_0 \) denotes the number of intermediate configurations.  

\begin{algorithm}[tpb]
\caption{Virtual Graph Generation (VGG)}
\KwIn{$Positions$}
\KwOut{$G_v$}

$G = Initialize(Positions)$\;
MakeValidConnect($G$)\;
$G_v = TraverseLeavesConfig(G)$\;
\textbf{return} $G_v$\;

\hrule
\textbf{function} MakeValidConnect($G$)\;
\ForEach{$Level$ in $\lceil$ Len($G.Block$) $\rceil$}{
    $Des = GetDescendants(G, Level)$\;
    $Pairs = MakePairs(Des)$\;
    \ForEach{$Pair$ in $Pairs$}{
        Connect($Pair$)\;
        $ValidPairs = Iscycle(G, Pair)$\;
        Disconnect($Pair$)\;
    }
}
$ValidPairs = NonRedundancy(ValidPairs)$\;
$ValidPairsBranch = NonConflict(ValidPairs)$\;
\ForEach{$ValidPairs$ in $ValidPairsBranch$}{
    \ForEach{$Pair$ in $ValidPairs$}{
        $NewG = Generate(Pair, G)$\;
    }
    $G$.AddChild($NewG$)\;
}
\end{algorithm}

To identify the intermediate graphs, we employ a method called the Bidirectional Isomorphism Tree (BIT), as illustrated in Fig.~\ref{fig:reconfiguration}.
Initially, the elements of $\mathcal{G}_{vs}$ and $\mathcal{G}_{vg}$ are respectively added as child nodes to two virtual root nodes. A random polyomino-shaped configuration $G_{ri}$ composed of multiple square modules is then generated using the polyominoes generator, followed by normalization to prevent duplication. Using $G_{r1}$ as an example, we demonstrate the expansion of BIT. Through the VGG algorithm, a corresponding set of virtual graphs $\mathcal{G}_{vr1}$ is obtained. Suppose there exists a graph $G_{vr1}^2 \in \mathcal{G}_{vr1}$ that is isomorphic to $G_{vs}^1 \in \mathcal{G}_{vs}$ (i.e., $G_{vr1}^2 \cong G_{vs}^1$). In this case, all elements except $G_{vr1}^2$ are excluded, and the remaining subset is denoted as $\tilde{\mathcal{G}}_{vr1}$, which is then added as the child node of $G_{vs}^1$ in the BIT. 

\subsection{Dependence-based Reverse Tree}

The generalized topological reconfiguration sequence, represented by Eq.~\eqref{eq4}, is the output of the BIT algorithm. This sequence consists of multiple action sets, each composed of VGG, Mapping, and AntiVGG operations, and can be further divided into two subsets: a connection action set $S_{\text{con}}$ and a disconnection action set $S_{\text{discon}}$. 

\begin{algorithm}[t]
\caption{DRTree Algorithm}
\label{alg:drtree}
\KwIn{$S_{con}$, $S_{discon}$, $G_1$, $G_2$}
\KwOut{$S_{action}$}

$S_{action} = [\,]$, $round = 1$\;
$Dep = $ ComputeDependencies($S_{con}$, $S_{discon}$, $G_1$, $G_2$)\;
$T_{dep}$ = Init($Dep$, $S_{con}$)\;
\While{$S_{con} \neq \emptyset$ \textbf{or} $S_{discon} \neq \emptyset$}{
    $C_{con} = $ CollectCandidates($S_{con}$, $T_{dep}$, $round$)\;
    $C_{dis} = $ CollectCandidates($S_{discon}$, $T_{dep}$, $round$)\;
    
    \eIf{$C_{con} = \emptyset$ \textbf{and} $C_{dis} = \emptyset$}{
        $round = round + 1$\;
        \lIf{$round > 3$}{\textbf{return} failure}
    }{
        $a^* = $ SelectBestAction($C_{con}$, $C_{dis}$, $T_{dep}$)\;
        
        \If{$a^* \in S_{con}$ \textbf{and} $Dep[a^*] \cap S_{discon} \neq \emptyset$}{
            $d^* = $ SelectDependency($a^*$, $T_{dep}$, $S_{discon}$)\;
            $S_{action}$.append($d^*$), $S_{discon}$.remove($d^*$)\;
        }
        
        $S_{action}$.append$(a^*)$, $S_{con}$.remove$(a^*)$\;
        HandleReinsertion($a^*$, $S_{con}$, $S_{discon}$, $G_1$, $G_2$)\;
        $round = 1$\;
    }
}

\textbf{return} $S_{action}$\;
\end{algorithm}

However, it should be noted that $S_{\text{con}}$ and $S_{\text{discon}}$ are not executable action sequences — sequential execution of elements in $S_{\text{con}}$ or $S_{\text{discon}}$ would render some morphpivotings infeasible, since the establishment of the virtual graph violates constraint 2. Therefore, it is necessary to establish a proper ordering by interleaving actions in $S_{\text{con}}$ with those in $S_{\text{discon}}$, thereby yielding an executable action sequence.

\begin{figure*}[tbp]
\centerline{\includegraphics[width=\linewidth]{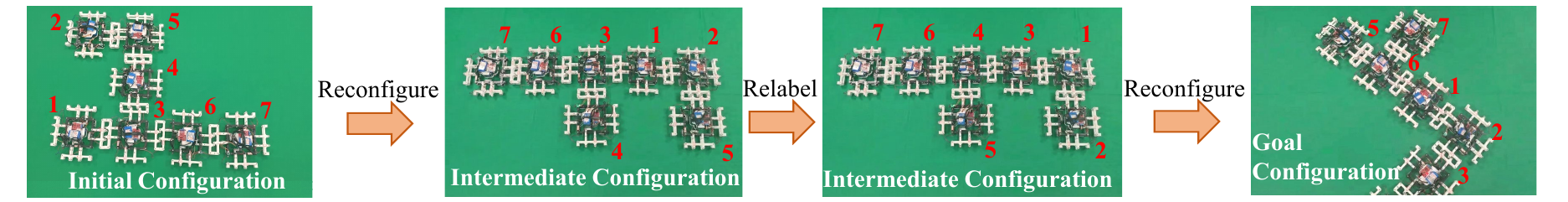}}
\caption{A reconfiguration example with one intermediate configuration. The initial configuration is first transformed into the intermediate configuration, which is then relabeled before being further transformed into the goal configuration.}
\label{fig:reconfigure2}
\end{figure*}

\begin{figure*}[tbp]
\centerline{\includegraphics[width=\linewidth]{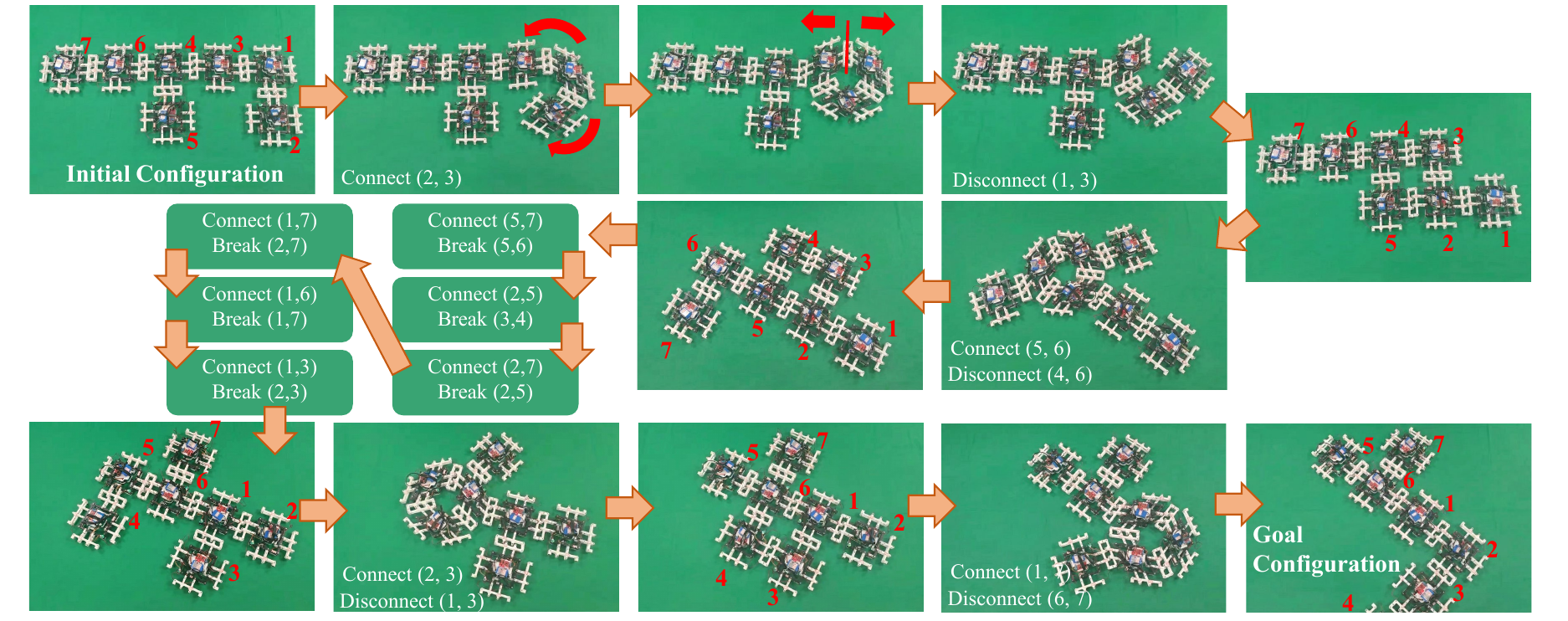}}
\caption{A reconfiguration example without intermediate configurations. The initial configuration is transformed into the goal configuration through ten pairs of actions, each consisting of one connection and one disconnection. Yellow arrows illustrate the transformation process of the robotic system, while red arrows indicate the directions of module connection and disconnection in each step.}
\label{fig:reconfigure}
\end{figure*}

We adopt a strategy to design the action sequence such that each element \( c_i \in S_{\mathrm{con}} \) is followed by a corresponding disconnection action to form a morphpivoting. This ensures that, upon the completion of all connection actions, every element in \( S_{\mathrm{discon}} \) has been incorporated into the sequence at its appropriate position.
Anchoring on $S_{\mathrm{con}}$, we propose the DRTree, as detailed in Algorithm~2, to integrate $S_{\mathrm{con}}$ and $S_{\mathrm{discon}}$.
The execution of $c_i$  must satisfy three constraints, particularly constraint 1, which requires all other edges in configuration graph forming a cycle with $c_i$ to be connected prior to executing $c_i$. These edges are defined as dependencies of $c_i$ and stored sequentially in $Dep$. Counterintuitively, these dependencies are assigned as child nodes of each $c_i$ in the DRTree structure. The initialization of the DRTree is completed in lines 3 of Algorithm 2, where elements from $S_{\text{con}}$ are inversely inserted into the tree according to $Dep$, with the last element serving as the root node. Line 4-20 employ a three-round progressive relaxation scheme to robustly interleave connection and disconnection actions under dependency constraints. In Round 1 (greedy), we admit only connections that have at least one removable dependency edge that is safe, and is present in the disconnection set, which reduces unnecessary backtracking. When Round 1 yields no feasible candidate, Round 2 (balanced) relaxes the rule by allowing dependencies that are either scheduled for disconnection or never needed again, improving feasibility while still controlling sequence growth. If both conservative rounds stall, Round 3 (fallback) accepts any connection whose dependencies are currently satisfied and resolves remaining conflicts through reinsertion/compensation, prioritizing completeness and guaranteed termination.

\section{Performance Evaluation and Experiment}
This section evaluates the computational efficiency of our proposed algorithm and introduces its implementation on physical hardware.

\subsection{Reconfiguration Demonstration}
Seven modules were fabricated for the experiments, the construction of which was briefly introduced in Section II. The specific control algorithm of the system is beyond the scope of this letter; the experiment mainly aims to emphasize and validate the feasibility of the reconfiguration sequence generated by our algorithm. 

Two random configurations were selected and provided as inputs to our algorithm. The results show that an intermediate configuration is required to accomplish the reconfiguration from the initial configuration to the goal configuration, as illustrated in Fig.~\ref{fig:reconfigure2}. The initial configuration is first transformed into the intermediate configuration, which is subsequently relabeled and then further transformed into the goal configuration. This procedure can also be regarded as two independent reconfiguration processes, namely from the initial to the intermediate configuration and from the intermediate to the goal configuration. In Fig.~\ref{fig:reconfigure}, the F-like configuration is again used as the initial configuration and provided to our algorithm, which further gives a step-by-step illustration of the reconfiguration process formed by alternating connection and disconnection actions.

\subsection{Performance Evaluation}
All computations were conducted on a desktop computer equipped with an Intel Core i9-12900KF (12th Gen, 3.20 GHz) processor and 32 GB of RAM. We compare our method against Bidirectional Rapidly-Exploring Random Tree (BiRRT)\cite{li2016birrtopt,odem2023rrt}, a widely used search-based planner for motion planning. To accommodate the motion constraints of our system, we modify BiRRT such that morphpivoting is treated as an indivisible primitive motion, rather than being decomposed into physically sub-operations. Furthermore, the distance metric between tree nodes is defined using the Manhattan distance over module groups.

\begin{table}[t]
\centering
\footnotesize
\caption{Algorithm Performance Comparison}
\begin{tabular}{
    c
    >{\centering\arraybackslash}p{1.3cm}
    >{\centering\arraybackslash}p{1.5cm}
    >{\centering\arraybackslash}p{1.3cm}
}
\toprule
\textbf{Algorithm} &
\makecell{\textbf{Success}\\\textbf{Ratio}} &
\makecell{\textbf{Avg. Min.}\\\textbf{Steps}} &
\makecell{\textbf{Avg. Time}\\\textbf{(s)}} \\
\midrule
Modified BiRRT  & 46\%  & 6.06 & 8.54 \\
Ours            & 100\% & 26.41 & 1.75 \\
\bottomrule
\end{tabular}
\end{table}

\begin{figure}[tbp]
\centerline{\includegraphics[width=\linewidth]{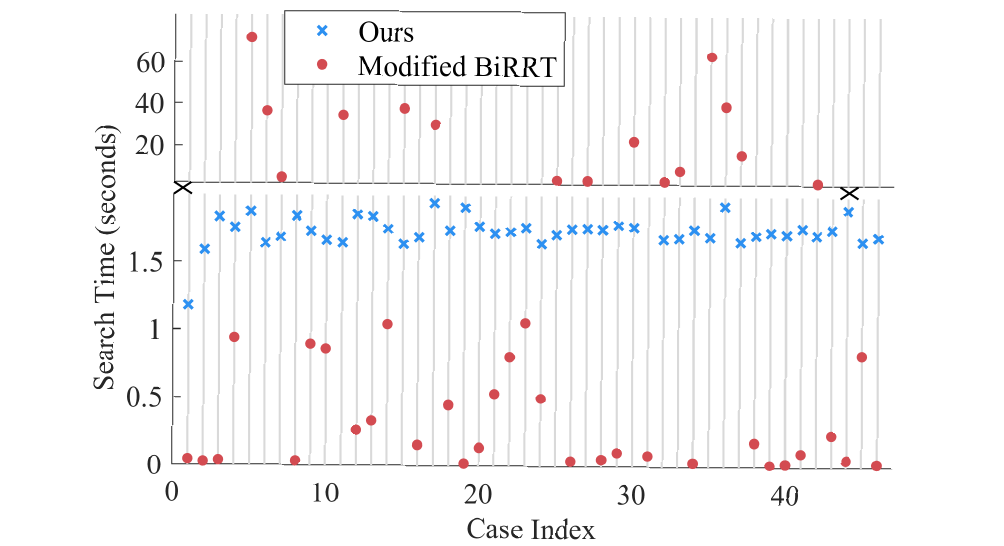}}
\caption{Comparison of the computation time for searching reconfiguration sequences between our algorithm and the modified BiRRT.}
\label{fig:comparison}
\end{figure}

Given that modular systems comprising seven or more units are theoretically capable of mutual reconfiguration between any two configurations (excluding strictly linear topologies), we evaluated the proposed method on a seven-module system. Both initial and target configurations were synthesized using a polyomino-based generator\cite{golomb1994}. To evaluate performance, we randomly generated 100 pairs of start and goal configurations, with each experimental trial repeated 20 times.

The aggregate results are presented in Table I. Compared to the baseline, our method demonstrates superior computational efficiency and ensures a high success rate, though it presents a trade-off regarding the optimality of the generated reconfiguration sequences. Specifically, to facilitate a direct comparison, we isolated the 46 cases where the modified BiRRT successfully converged; the performance contrast on these instances is illustrated in Fig. \ref{fig:comparison}.

The experimental results highlight a marked disparity in performance stability. While the modified BiRRT efficiently solves low-complexity reconfiguration tasks, its stochastic nature leads to significant struggles in complex scenarios, frequently resulting in excessive search times or convergence failures (exceeding the 1,000-iteration cutoff). In contrast, the proposed algorithm demonstrates superior robustness, maintaining consistent computation times and achieving a 100\% success rate across all trials, regardless of task complexity.

\section{Discussion of Reconfigurability}
Morphpivoting ensures robust docking but requires at least three participating modules, which prevents universal reconfiguration between arbitrary configurations. To determine if two configurations are mutually reachable (isotypic), we define an isotypic signature, \textit{Maximum Loop Size} ($\mathcal{S}$): the size of the largest loop achievable by a configuration $C$ or its reachable variants.

As shown in Fig.~\ref{fig:Configuration_Graph}(a), the canonical configuration can form a pentagonal loop ($\mathcal{S}=5$) but cannot reach a hexagonal one due to the indispensability of the auxiliary module (labeled 1). Thus, it belongs to the component $\mathcal{C}_6^5$. 

Based on the above definition, each configuration possesses a unique value of $\mathcal{S}$, 
and configurations sharing the same $\mathcal{S}$ are considered isotypic. 
Accordingly, the entire configuration space $\mathcal{C}_n$ can be partitioned into multiple connectivity components 
\(\mathcal{C}_n^{\mathcal{S}}\), within which all configurations are isotypic. 
Each component is uniquely characterized by its corresponding $\mathcal{S}$.

When the configuration space is relatively small, it is feasible to enumerate all possible canonical configurations 
and classify them to exhaustively determine all possible values of $\mathcal{S}$. 
For example,
\begin{equation}
\begin{aligned}
\mathcal{C}_1 &= \mathcal{C}_1^0, \quad 
\mathcal{C}_2 = \mathcal{C}_2^0, \\
\mathcal{C}_3 &= \{\mathcal{C}_3^0\}, \quad 
\mathcal{C}_4 = \{\mathcal{C}_4^0, \mathcal{C}_4^3\}, \\
\mathcal{C}_5 &= \{\mathcal{C}_5^0, \mathcal{C}_5^3, \mathcal{C}_5^4\}, \quad 
\mathcal{C}_6 = \{\mathcal{C}_6^0, \mathcal{C}_6^3, \mathcal{C}_6^5\}.
\end{aligned}
\label{eq:configuration_space}
\end{equation}

When the number of modules increases, the following theorem can be established.

\textbf{Theorem 1:} When \(n \geq 7\), $\mathcal{S}$ has only two values: 0 and \(n-1\), that is, $\mathcal{C}_n = \{\mathcal{C}^{n-1}_n, \mathcal{C}^0_n \}$.

\begin{figure}[tbp]
\centerline{\includegraphics[width=\linewidth]{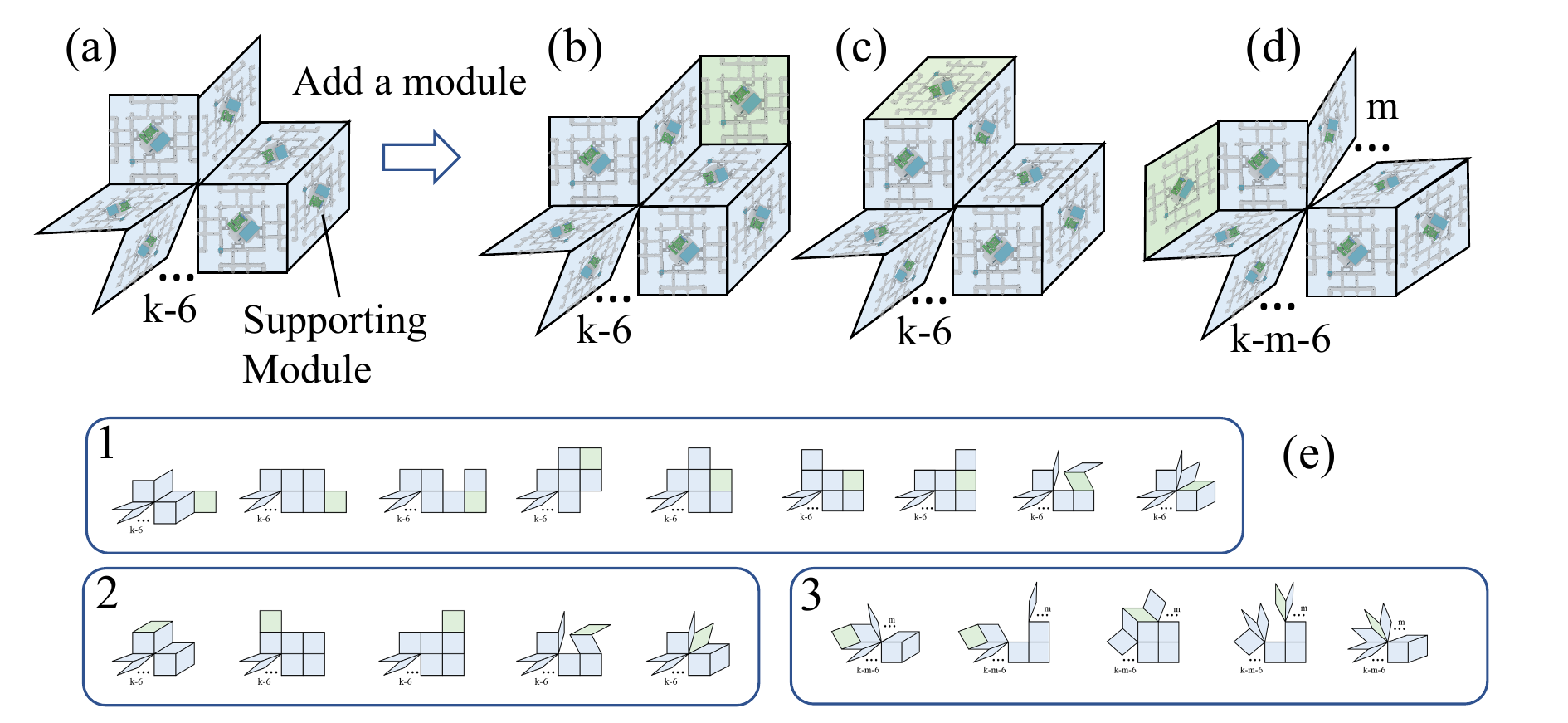}}
\caption{Proof of the non-existence of the set $\mathcal{C}_n^{k-1}$. 
(a) The subconfiguration composed of a $k$-loop and an auxiliary module. 
(b) Adding a module at the first gap in the counter-clockwise direction. 
(c) Adding a module at the second gap in the counter-clockwise direction. 
(d) Adding a module at the $(m+2)$-th gap in the counter-clockwise direction. 
(e) Transformation process into a $(k+1)$-loop for cases (b), (c), and (d).}
\label{fig:reconfigurability}
\end{figure}

\textbf{Proof:}  
Values of 1 and 2 are invalid for $\mathcal{S}$, as forming a loop requires at least three modules. Furthermore, for $n \geq 7$, $\mathcal{S}$ cannot equal 3, since the constructed virtual graph invariably contains a tetragonal loop. Assume the existence of \( \mathcal{C}_n^{k} \) for \( k = 4, 5, \ldots, n-2 \). Then, for any configuration \(C \in \mathcal{C}_n^{k}\), it must be possible to reconfigure into a configuration, which can be decomposed into a subconfiguration consisting of a \(k\)-loop and an auxiliary module, shown as Fig.~\ref{fig:reconfigurability}(a), along with \(n-k-1\) remaining modules. By selecting one module from the remaining modules and adding it to the subconfiguration, three possible scenarios arise, as shown in Fig.~\ref{fig:reconfigurability}(b)--(d), and their corresponding reconfiguration processes into a $(k+1)$-loop are illustrated in Fig.~\ref{fig:reconfigurability}(e), labeled as 1, 2, and 3. The formation of a $(k+1)$-loop contradicts the assumption that $C \in \mathcal{C}_n^{k}$. This implies that no such configuration exists.

Based on Theorem 1, we derive the following corollary, which guarantees the existence of a topological reconfiguration sequence between any two configurations:  

\textbf{Corollary 1:}  
For \(n \geq 7\), any two configurations (excluding linear configurations) are isotypic.  

\textbf{Proof:}  
Any non-linear configuration must contain three modules forming a corner (L-shape), which generates a 3-cycle. Consequently, $\mathcal{S}$ is at least 3. By Theorem 1, when \(n \geq 7\), the $\mathcal{S}$ can only be \(n-1\). Therefore, all configurations except the linear configuration belong to \(\mathcal{C}^{n-1}_n\).

\section{Conclusion and Future Work}
In this letter, we proposed a novel self-reconfiguration planning algorithm for deformable quadrilateral MSRRs. The core of our approach involves a two-stage strategy: first, generating a set of unordered reconfiguration actions by leveraging virtual graph construction and isomorphic mapping search; and second, organizing these actions via a novel method termed the DRTree. Physical experiments demonstrated the practical feasibility of the generated reconfiguration sequences. Moreover, comparisons with the modified BiRRT algorithm highlighted the proposed method's advantages in terms of stability and planning efficiency. Additionally, we provided a theoretical proof establishing the reconfigurability between configurations under specific topological constraints.

Our future work will focus on improving the optimality of reconfiguration sequences. Although the proposed approach generates actions rapidly, it does not guarantee a globally minimal solution. While our method consistently identifies valid paths, it may yield longer reconfiguration sequences in certain simple scenarios. Additionally, we plan to extend our algorithm to other polygonal robot systems and validate these generalizations on physical hardware platforms.

\addtolength{\textheight}{-12cm}   


\bibliographystyle{IEEEtran}
\bibliography{references}

\end{document}